\documentclass[letterpaper, 10pt, conference]{ieeeconf}      % Use this line for a4 paper

\IEEEoverridecommandlockouts                              % This command is only needed if 
\usepackage{cite}
\usepackage{amsmath,amssymb,amsfonts}
\usepackage{algorithmic}
\usepackage{graphicx}
\usepackage{textcomp}
\usepackage{xcolor}
\usepackage{blindtext}
\usepackage[hyphens]{url}
\usepackage{hyperref}
\usepackage{float}
\usepackage{acro}
\usepackage[labelformat=simple]{subcaption}
\usepackage{siunitx}

\definecolor{twblue}{HTML}{1f497d}
\definecolor{structure}{HTML}{1f497d}
\definecolor{blindtext}{HTML}{D3D3D3}
\definecolor{review}{HTML}{f25e5e}

\DeclareAcronym{ros}{short = ROS, long = Robot Operating System}
\DeclareAcronym{slam}{short = SLAM, long = simultaneous localisation and mapping}
\DeclareAcronym{lidar}{short = LiDAR, long = light detection and ranging}
\DeclareAcronym{imu}{short = IMU, long = inertial measurement unit}
\DeclareAcronym{gpr}{short = GPR, long = Gaussian process regression}
\DeclareAcronym{gp}{short = GP, long = Gaussian process}
\DeclareAcronym{usar}{short = USAR, long = Urban search and rescue}
\DeclareAcronym{sar}{short = SAR, long = search and rescue}
\DeclareAcronym{tcp}{short = TCP, long = tool center point}
\DeclareAcronym{nbc}{short = NBC, long = {nuclear, biological, and chemical}}
\DeclareAcronym{enrich}{short = EnRicH, long = European Robotics Hackathon}
\DeclareAcronym{ugv}{short = UGV, long = Unmanned ground vehicle, long-plural =s}
\DeclareAcronym{uav}{short = UAV, long = unmanned areal vehicle}
\DeclareAcronym{dof}{short = DoF, long = degree of freedom}
\DeclareAcronym{abc}{short = ABC, long = {atomic, biological, and chemical}}
\DeclareAcronym{cbrn}{short = CBRN, long = {chemical, biological, radiological, and nuclear}}

\title{\LARGE \bf
UGV-CBRN: An Unmanned Ground Vehicle for \\ Chemical, Biological, Radiological, and Nuclear Disaster Response
}

\author{Simon Schwaiger$^{*1,2}$, Lucas Muster$^{*1,3}$, Georg Novotny$^{1}$, Michael Schebek$^{1}$,\\Wilfried Wöber$^{1}$, Stefan Thalhammer$^{1}$ and Christoph Böhm$^{1}$% <-this % stops a space
\thanks{*Equal contribution}
\thanks{This work was supported by the Austrian Research Promotion Agency (FFG) through the research project UGV-ABC-Probe (FFG project Call 2020) and the Austrian Armed Forces.}% <-this % stops a space
\thanks{$^{1}$Simon Schwaiger, Lucas Muster, Georg Novotny, Michael Schebek, Wilfried Wöber, Stefan Thalhammer and Christoph Böhm are with the University of Applied Sciences Technikum Wien, Faculty of Industrial Engineering, 1200 Vienna, Austria {\tt\small schwaige@technikum-wien.at}, {\tt\small muster@technikum-wien.at}}%
\thanks{$^{2}$Simon Schwaiger is with the Graz University of Technology, Faculty of Computer Science and Biomedical Engineering, Institute of Software Technology, Inffeldgasse 16b/II, 8010 Graz, Austria}%
\thanks{$^{3}$Lucas Muster is with the University of Natural Resources and Life Sciences, Department of Biotechnology, Institute for Computational Biology, Muthgasse 18, 1190 Vienna, Austria}%
}

\newcommand{\figref}[1]{Fig.~\ref{fig:#1}}
\newcommand{\secref}[1]{Sec.~\ref{sec:#1}}

\newcommand{\equref}[1]{Eq.~(\ref{eq:#1})}

\begin{document}

\maketitle
\thispagestyle{empty}
\pagestyle{empty}

%%%%%%%%%%%%%%%%%%%%%%%%%%%%%%%%%%%%%%%%%%%%%%%%%%%%%%%%%%%%%%%%%%%%%%%%%%%%%%%%
\begin{abstract}

Robotic search and rescue (SAR) supports response teams by accelerating disaster assessment and by keeping operators away from hazardous environments.
In the event of a chemical, biological, radiological, and nuclear (CBRN) disaster, robots are deployed to identify and locate radiation sources. Human responders then assess the situation and neutralize the danger.
The presented system takes a step toward enhanced integration of robots into SAR teams. 
Integrating autonomous radiation mapping with semi-autonomous substance sampling and online analysis of the CBRN threat lets the human operator localize and assess the threat from a safe distance.
Two LiDARs, an IMU, and a Geiger counter are used for mapping the surrounding area and localizing potential radiation sources.
A mobile manipulator with six Degrees of Freedom manipulates valves and samples substances that are analyzed by an onboard Raman spectrometer.
The human operator monitors the mission's progression from a remote location defining target locations and directing the semi-autonomous manipulation processes. 
Diverse recovery behaviours aid robot deployment, system state monitoring, as well as recovery of hard- and software.
Field tests showcase the capabilities of the presented system during trials at the CBRN disaster response challenge \acl{enrich} (\acs{enrich}).
We provide recorded sensor data and implemented software through a GitHub repository: \href{https://github.com/TW-Robotics/search-and-rescue-robot-IROS2024}{\url{https://github.com/TW-Robotics/search-and-rescue-robot-2024}}.

\end{abstract}
%%%%%%%%%%%%%%%%%%%%%%%%%%%%%%%%%%%%%%%%%%%%%%%%%%%%%%%%%%%%%%%%%%%%%%%%%%%%%%%%
\section{INTRODUCTION} \label{sec:introduction}

% Problem
Robotic \ac{sar} systems aid disaster emergency teams by keeping operators and humans away from danger and by reducing the time needed for disaster assessment~\cite{murphy2004trial,Queralta2020}.
\acp{ugv} are the prevalent choice due to their high payloads and versatility when it comes to sensors, operation time, and robustness~\cite{Berns17}.

Especially radiation mapping is important to ensure operators distance to hazardous materials and to accelerate relief in \ac{cbrn} disasters, where exposure time is a critical factor~\cite{ohno2011robotic, rosas2020autonomous,Groves21,GPR:radMap:RESCUER,abd2020mobile,abd2022coverage,nouri2023carma}.
\ac{usar} robots support emergency teams by measuring radiation autonomously or via teleoperation~\cite{ohno2011robotic,rosas2020autonomous}.
Autonomous localization and mapping of radiation is performed using priori maps or by simultaneously exploring and mapping the environment~\cite{abd2020mobile,abd2022coverage,nouri2023carma}.
Additionally, strategies for avoiding radiation sources during autonomous exploration and the self-contamination of the robot after exposure have been proposed to limit the spread of contaminant~\cite{Groves21,GPR:radMap:RESCUER}.
However, supporting \ac{cbrn} missions with robotic environment interaction and the robot's ability to manipulate and analyse \ac{cbrn} contaminants, accelerates disaster site clearance and increases the protection of human operators by reducing contaminant exposure times and intensity.

\begin{figure}[t]
    \centering
    \includegraphics[width=0.45\textwidth]{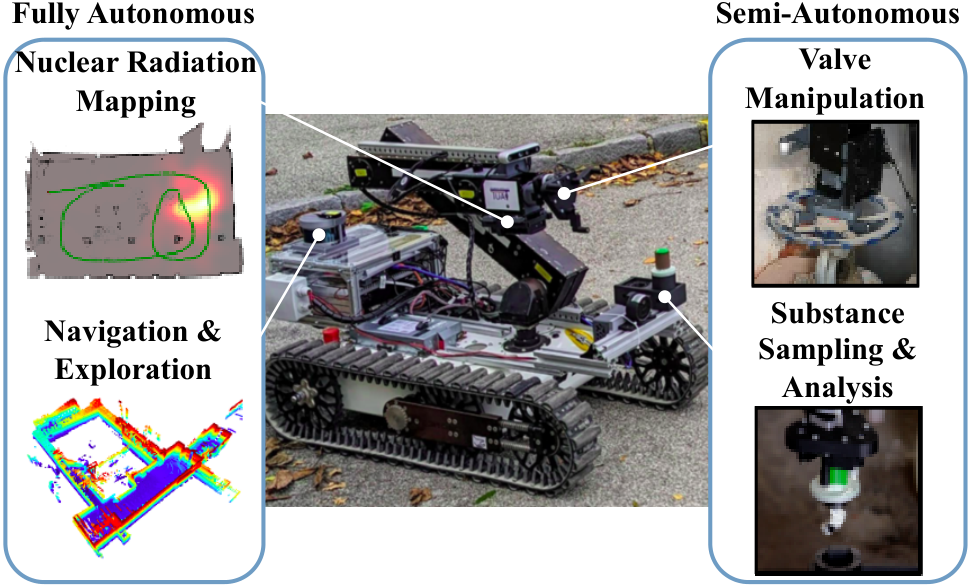}
    \caption{\textbf{UGV-CBRN} A robotic system for assisting disaster response through geometry and radiation mapping, substance sampling and analysis, as well as valve manipulation.}
    \label{fig:task-overview}
\end{figure}

This manuscript presents UGV-CBRN, an \textbf{U}nmanned \textbf{G}round \textbf{V}ehicle for \textbf{C}hemical, \textbf{B}iological, \textbf{R}adiological, and \textbf{N}uclear disaster response, illustrated in~\figref{task-overview}.
Autonomous navigation, exploration, and geometrical and radiation mapping, are integrated with semi-autonomous valve closing and substance sampling and analysis.
Crucial tasks that require human intervention to make the decision are semi-autonomous, i.e. valve closing and substance sampling. 
These tasks are performed with a prototyped 2-in-1 end-effector. 
Multiple system recovery behaviors complement system monitoring to increase resilience to failures.
UGV-CBRN has been designed and implemented to support the Austrian Armed Forces for solving their challenges in disaster response, and was tested at the \acl{enrich} (\acs{enrich})\footnote{\url{https://enrich.european-robotics.eu/}}, in 2023.

%% Contribution
In summary our contributions are:

\begin{itemize}
    \item an unmanned ground vehicle for chemical, biological, radiological, and nuclear disaster response. The system performs integrated autonomous geometrical and radiation mapping, navigation and exploration, as well as manipulation for sampling substances and stopping contaminant leakage.
    \item routines for flexible semi-autonomy and error recovery. Enabling autonomy for mapping, navigation, and exploration, and semi-autonomous teleoperation for those that necessitate a considerable degree of responsibility, i.e. substance sampling and stopping contaminant leakage.
    \item field trials showcasing the individual capabilities in a \ac{cbrn} response challenge and a disaster relief training area.
\end{itemize}

The remainder of this paper is structured as follows. \secref{related_work} presents recent advances in autonomous \ac{sar} systems. \secref{method} provides a methodological overview of the \ac{sar} robot and the fulfilled tasks, while \secref{implementation_detail} discusses robot hardware and software. \secref{experiments} describes the robot's evaluation in three scenarios and \secref{conclusions} summarizes our research and provides an outlook on future work.

\section{RELATED WORK} \label{sec:related_work}

The first documented use of robots in urban search and rescue was at securing the World Trade Center disaster in $2001$~\cite{murphy2004trial}.
Since then large progress has been made for supporting human operators with teleoperated, semi-, and full autonomous robotic systems.
Common aspects of \ac{sar} robotics are exploring, mapping and monitoring disaster zones with solitary~\cite{rosas2020autonomous,arabboev2021development,nosirov2020specially} and multi-robot systems~\cite{Berns17,wang2023development,zhao2017search}, localizing and mapping radiation~\cite{ohno2011robotic,Groves21,abd2022coverage,nouri2023carma,GPR:radMap:RESCUER}, and mobile manipulation for debris and obstacle removal~\cite{Berns17,engemann2020omnivil,ramasubramanian2021operator,wang2023development}. % a system for exploring and mapping disaster areas is presented.

\textbf{Dealing with Radiation in SAR}
The authors of~\cite{ohno2011robotic} present a teleoperated robot to remotely measure radiation.
The system of~\cite{Groves21} is designed to measure and avoid areas with high radiation.
In~\cite{GPR:radMap:RESCUER} a system for decontamination after exposure to harmful substances is presented.
The authors of~\cite{abd2020mobile,abd2022coverage} present systems for mapping radiation given an a priori map.
Another system for mapping radiation is presented in~\cite{nouri2023carma}, where the authors provide functionalities for densely mapping the environment.
A promising strategy for obtaining radiation maps is estimating the posterior predictive distribution of a Poisson distribution~\cite{GPR:radMap:Basics}.
In a follow up work a radiation mapping pipeline optimized for autonomous mobile robots is proposed~\cite{GPR:radMap:Nature}. 
The authors apply non-parametric \ac{gpr}~\cite{GPR:Rasmussen} to map the radiation using Geiger counts. 

\textbf{Mobile Manipulation in SAR}
Mobile manipulation in SAR is usually achieved by deploying multi-robot systems.
In~\cite{Berns17} two mobile ground vehicles, a mobile base for removing debris and obstacles and a smaller one for mapping, are presented.
The authors of~\cite{zhao2017search} present two robots where the smaller one provides the possibility for tele-operated manipulation with a three \ac{dof} robotic manipulator.
Similary,~\cite{wang2023development} present also a dual-robot system with a six \ac{dof} manipulator with a two-finger gripper for opening doors and clearing roadblocks

Possessing manipulation capabilities is common for \ac{sar} robots; however, there is a absence of integration with \ac{cbrn} disaster response. 
The combination of these capabilities is crucial for the expeditious clearance of disaster sites and for protecting human operators from harm.
To the best of our knowledge the presented system is the first of its kind to provide radiation mapping, substance sampling, and online substance analysis for \ac{cbrn} scenarios.

\begin{figure*}[tb!]
    \centering
    \vspace{2mm}
    \includegraphics[width=.88\textwidth]{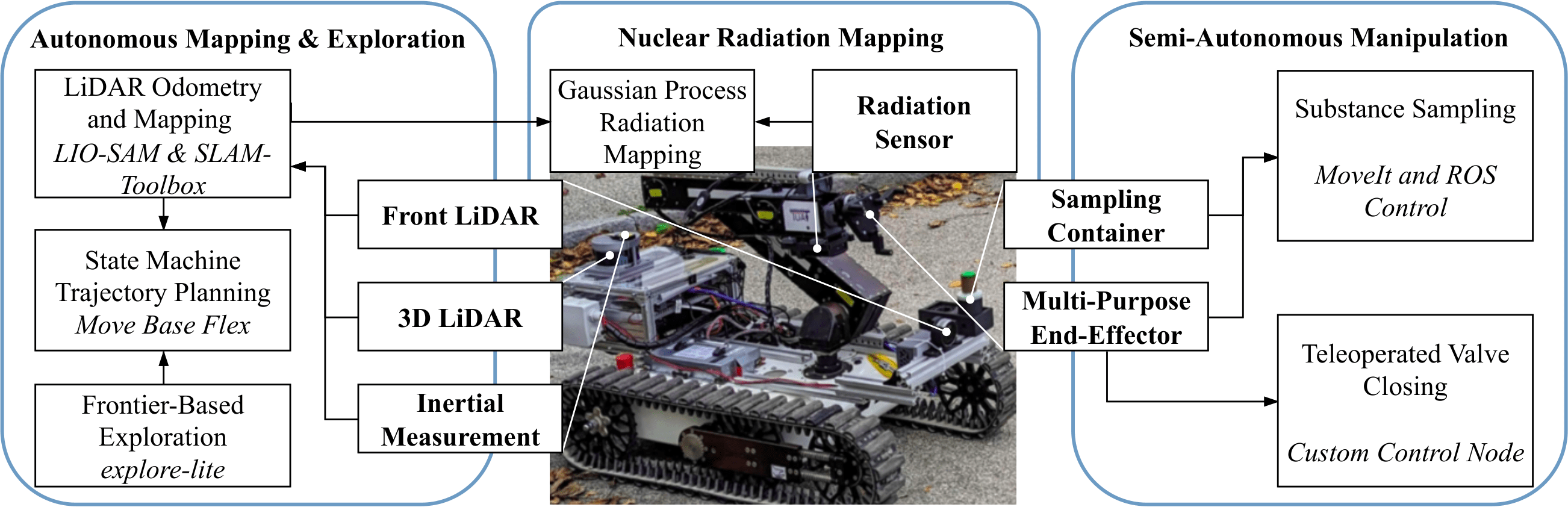}
    \caption{\textbf{UGV-CBRN: Integrated system overview} The proposed system aims to provide 2D and 3D mapping, radiation mapping, autonomous navigation and manipulation capabilities (blue rounded border) for \ac{sar} missions. The figure depicts the data flow between functional units, sensors and the multi-purpose end-effector.}
    \label{fig:system_overview}
\end{figure*}

\section{Integrated Robotic System} 
\label{sec:method}

This sections details the presented \textbf{U}nmanned \textbf{G}round \textbf{V}ehicle with \textbf{C}hemical, \textbf{B}iological, and \textbf{R}adiological, and \textbf{N}uclear analysis capabilities.
\figref{system_overview} presents an overview of the proposed system.
The system is based on a pre-production version of the Taurob Tracker\footnote{\url{www.offshore-technology.com/news/ogtc-offshore-robot-development/}} supplemented with a custom gripper that provides two additional \ac{dof} to the arm.
The front facing Realsense L515\footnote{\url{www.intel.com/content/www/us/en/products/sku/201775/intel-realsense-lidar-camera-l515/specifications.html}},
a \ac{lidar}\footnote{\url{ouster.com/products/hardware/vlp-16}} sensor, and an \ac{imu}\footnote{LORD 3DMGX5-GNSS/INS} capture data for geometric mapping, navigation and exploration of disaster areas. 
A radiation detector\footnote{\url{www.cbrnetechindex.com/Print/4277/flir-systems-inc/identifinder-r300}} measures radiation.
A multi-purpose end-effector with a Realsense D455\footnote{\url{www.intelrealsense.com/depth-camera-d455/}} manipulates valves and samples \ac{cbrn} substances, which are analysed online with a Raman spectrometer\footnote{\url{www.fishersci.com/shop/products/thermo-scientific-firstdefender-rmx-handheld-chemical-identification-5/p-4006497}}.
The built-in industrial PC features an \textit{Intel i7 7700T} CPU, \textit{Nvidia GTX 1050 Ti} GPU and 32 GB of system memory.
An \textit{Nvidia Jetson Xavier NX 16GB} single board computer interfaces with the D455 and the front \ac{lidar}.

The software stack is operated by a synergistic distribution of \ac{ros}~\cite{ROS} nodes running in parallel in Docker~\cite{Novotny2023}.
UGV-CBRN is designed for seamless transition between semi- and autonomous operation. 
While the system provides a high degree of autonomy, the operator verifies and refines end-effector poses for manipulation. 

% Software Setup
\subsection{Mapping and Exploration}
\label{sec:2d3dmapping}

While the use of a 3D mapper is preferred over its 2D counterpart, encountered challenges w.r.t. the onboard PC's computational performance during trials have led to a 2D mapper being deployed alongside the 3D mapper.
Autonomous navigation is backed by the more efficient 2D \ac{slam}, while the 3D mapper provides the robot's operator with a more detailed representation of the robot's surrounding as compared to a 2D map.
To reduce system latency when other on-board systems, such as the manipulation system, are active, the 3D mapper can be disabled without affecting navigation.

The 2D mapper was selected based on its wide adoption as the by default supported mapper in \ac{ros} 2 \cite{Macenski2021SLAMToolbox}. 
The system employs odometry and \ac{lidar} measurements, while the 3D mapper relies on the combination of \ac{lidar} and \ac{imu}, which is more robust than \ac{lidar}-only 3D mapping, particularly when the robot exhibits sudden jerking or rotation, as observed during field testing~\cite{Lee2024LidarOdometrySurvey}. Due to its performance during field trials and the compatibility with available hardware, the implementation of~\cite{liosam2020shan} is used.

\textbf{Autonomous Navigation}
Autonomous navigation is achieved by planning a trajectory from the current to the goal pose provided by frontier-based exploration~\cite{Hoerner2016}.
For planning trajectories to goal poses layered costmaps~\cite{Lu2014LayeredCostmaps} have been established as the standard approach for mobile robot trajectory planning with ROS~\cite{Puetz2018MoveBaseFlex}.

Field tests showed that tight indoors, common in \ac{usar}, and track slippage can lead to the robot being unable to accurately follow the planned trajectory.
To circumvent this issue, costmap-based trajectory planning \cite{Lu2014LayeredCostmaps} is reinforced by a finite state machine~\cite{Puetz2018MoveBaseFlex} triggering complementary recovery behaviors. 
In order to pass through narrow doors it has been shown to be crucial to reduce the robot's padding in the costmap to \SI{50}{\mm}.
This, however, increases the risk of collisions. 
In order to counteract this, additional collision detection and recovery is implemented.
Collisions are detected if the \ac{lidar} measures obstacles to be closer than \SI{0.1}{\m} to the robot's footprint.
If a collision is detected, the robot backtracks \SI{0.3}{\m} along the travelled path, clears the costmap and attempts re-planning.
If it still fails after the third attempt, the robot aborts the goal.

\subsection{Nuclear Radiation Mapping}

Our implementation is inspired by the work of~\cite{GPR:radMap:Nature} to obtain a 2D map.
The mobile robot's current pose is combined with the counts of the radiation detector.
At each time step $t$ the gathered radiation count data $c_t$ is synchronized with the robot's pose $x_t$ estimated by the 2D mapper~\cite{Macenski2021SLAMToolbox}.
The training data $\mathcal{D}$ describes the gathered poses and count data.

The map is generated by defining the posterior predictive distribution $p(c_{new}|x_{new},x_{1:t},c_{1:t})=p(c_{new}|x_{new},\mathcal{D})$ 
relying on \ac{gpr} with radial base function kernels \cite{GPR:Rasmussen}.
Hence, for each location $(i,j)$ in the map $m$, the radiation is estimated using \equref{rad_mapping}, where the functions $GP_\mu(.)$ and $GP_\Sigma(.)$ describe the prediction of the \ac{gp}.
\begin{multline}\label{eq:rad_mapping}
    \forall{i,j} \in m, p(c_{i,j}|x_{i,j},\mathcal{D})\approx \\
    \mathcal{N}\left(GP_\mu(x_{i,j}, \mathcal{D}),GP_\Sigma(x_{i,j},\mathcal{D})\right)
\end{multline}

Alignment of the geometric and the radiation map is performed by the operator to reduce system in the case that no radiation source is anticipated. 

% grippin'
\subsection{Manipulation, Sampling, and Analysis}
\begin{figure}[t!]
    \centering
    \vspace{2mm}
    \includegraphics[width=0.92\columnwidth]{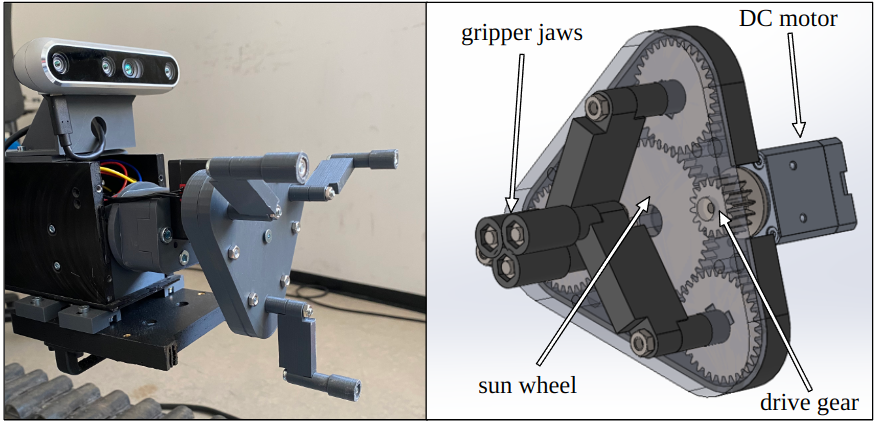}
    \caption{\textbf{Multi-purpose end-effector} The left part of the figure shows a real image of the end-effector and the right part shows the schematics and integral components.}
    \label{fig:gripper_schematic}
\end{figure}

For manipulation tasks UGV-CBRN is equipped with an industrial robotic arm with four \ac{dof}. 
The custom multi-purpose end-effector, \figref{gripper_schematic} shows the manipulation system on the left and its schematics on the right.

\textbf{Valve Manipulation}
Assuming centered placement of the \ac{tcp} along a valve's rotation axis, the end-effector manipulates valves of up to \SI{160}{\mm} in diameter with an external grip.
Valves with a diameter larger than \SI{160}{\mm} are handled with an internal grip.
Valves are opened and closes by rotating the end-effector base. 

\textbf{Substance Sampling}
The sampling mechanism is designed to initialize the analysis of hazardous substances with minimal human intervention. 
The end-effector is capable of grasping a specially designed probe, which is used to collect samples from potentially contaminated surfaces.
This probe is constructed with a cylindrical interface designed for maximum error tolerance when gripping it, and allowing precise control during the sampling process.

\textbf{Substance Analysis}
At the end of the probe, a glass wool swab is attached, designed to efficiently absorb substances for latter analysis.
Probes containing sampled substances are autonomously positioned in the designated storage tray on the robot, where the sample is deposited for latter analysis. 
The tray holds up to two probes.
One additional probe slot is present at the back of the robot, where the sampled substance is directly analysed on site with a handheld Raman spectrometer for direct chemical substance analysis.

% UGV-ABC OS
\subsection{Software Operating System} \label{sec:implementation_detail}

Our system uses a synergistic integration of Docker containers\footnote{\url{www.docker.com/}} to encapsulate software packages, systemd service management\footnote{\url{systemd.io/}}, and the Cockpit monitoring tool\footnote{\url{cockpit-project.org/}}~\cite{Novotny2023}.
This allows each software package's diverse dependencies to be managed individually. 
It enables to set up software in a portable and hardware-agnostic manner, while selectively allowing communication between software components when needed.
Systemd, a service management system, automates and manages the lifecycle of Docker containers in groups that represent the main stages (sensing, perception, mapping, exploration, navigation) of the \ac{sar} robot's data processing pipeline.
Systemd ensures that these groups are reliably started, stopped, and restarted as needed based on whether or not the robot's core drivers are running.

\textbf{Cockpit Monitoring}
The cockpit monitoring tool provides a comprehensive view of the system status, including the health and activity of both Docker containers and systemd services, facilitating efficient monitoring by human operators.
By immediate identification and troubleshooting of potential problems, the \ac{sar} robot's continuous availability is guaranteed.
Cockpit monitoring allows Docker containers, systemd services, and \ac{ros} nodes to be started, stopped and restarted manually using the graphical user interface, without requiring system restart. 
This functionality is not native in \ac{ros} when using the robot control pipeline.
Its utility has been proven crucial for successful mission completion.

Systemd services start orchestration of the robot processing pipeline components, regardless of the availability of a roscore. Therefore, the roscore needs to be monitored to determine if the system is operational.
The monitoring is achieved by repeatedly querying the output of \ac{ros} command line tools, however, this constant pinging of the roscore results in a significant computational load during system startup with a high number of monitored containers.
Dynamically adjusting the querying rate effectively minimizes this issue.
In our field trials the querying rate was set to \SI{0.2}{\hertz}.

\textbf{Missing Internet Connection during Deployment}
During deployment re-configuring parameters requires rebuilding the Docker containers, which in turn requires internet connection for installing the required dependencies.
Since an internet connection is typically unavailable during deployment, dependencies are declared in multi-stage Dockerfiles.
Local configuration changes hence only require partial rebuilds of containers cached by the Docker engine.
Internet is required to install base dependencies.
Which is done before copying any program files to the container.
Furthermore, containers have configuration files and \ac{ros} launch files mounted at runtime.
This approach allows configuration files to be changed without causing a container rebuild.

%%%%%%%%%%%%%%%%%%%%%%%%%%%%%
% SYSTEM OPERATION
\subsection{System Operation}

\begin{figure}[t!]
    \centering
    \vspace{2mm}
    \includegraphics[width=.86\columnwidth]{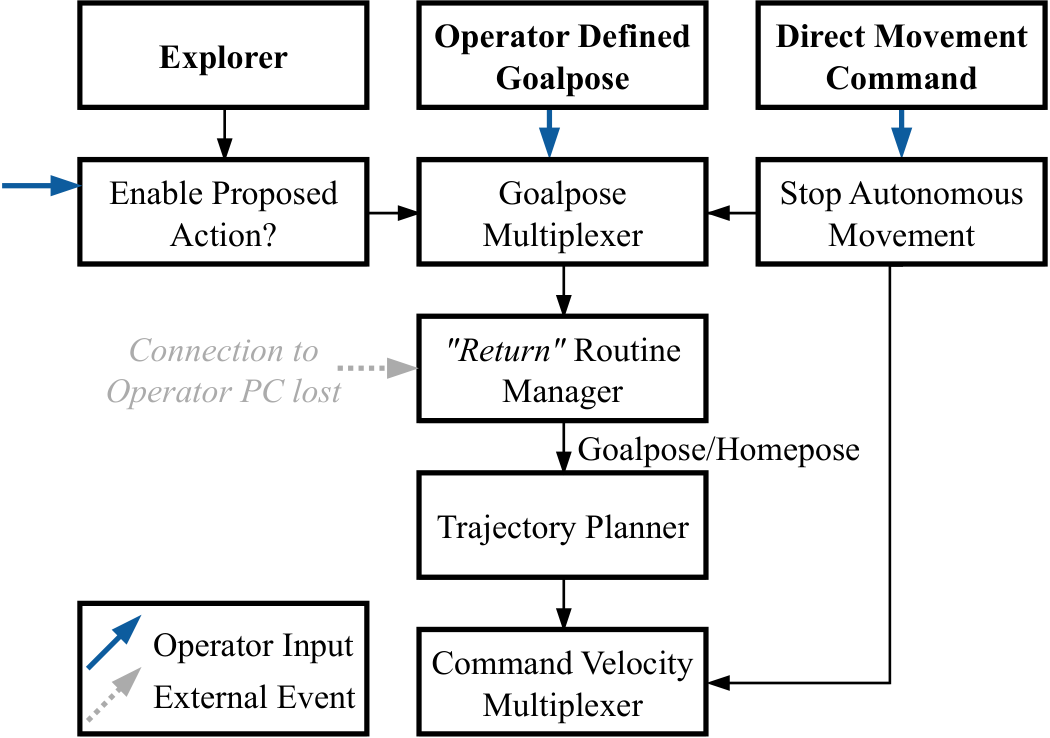}
    \caption{\textbf{UGV-CBRN: System operation} The proposed system switches between autonomous navigation and operator input based on predetermined priorities. Manual input is always preferred over autonomous behavior.}
    \label{fig:robot_behavior}
\end{figure}

UGV-CBRN operation is semi-autonomous and is supplemented with recovery behaviors.
Operation modes are switched based on priorities and the robot state.
This design philosophy has been established to be optimal for \ac{sar} missions~\cite{zhang2023earthshaker}.
Control signals given by the human operator are favored over control signals given by the exploration or trajectory planning modules.
This approach allows the robot's users to interrupt autonomous behavior by providing direct control signals or setting goal poses using the graphical user interface.
\figref{robot_behavior} presents the autonomous navigation stack's signal flow as well as interaction with the operator.

\begin{figure*}[!t]
    \centering
    \vspace{2mm}
    \includegraphics[width=0.94\textwidth]{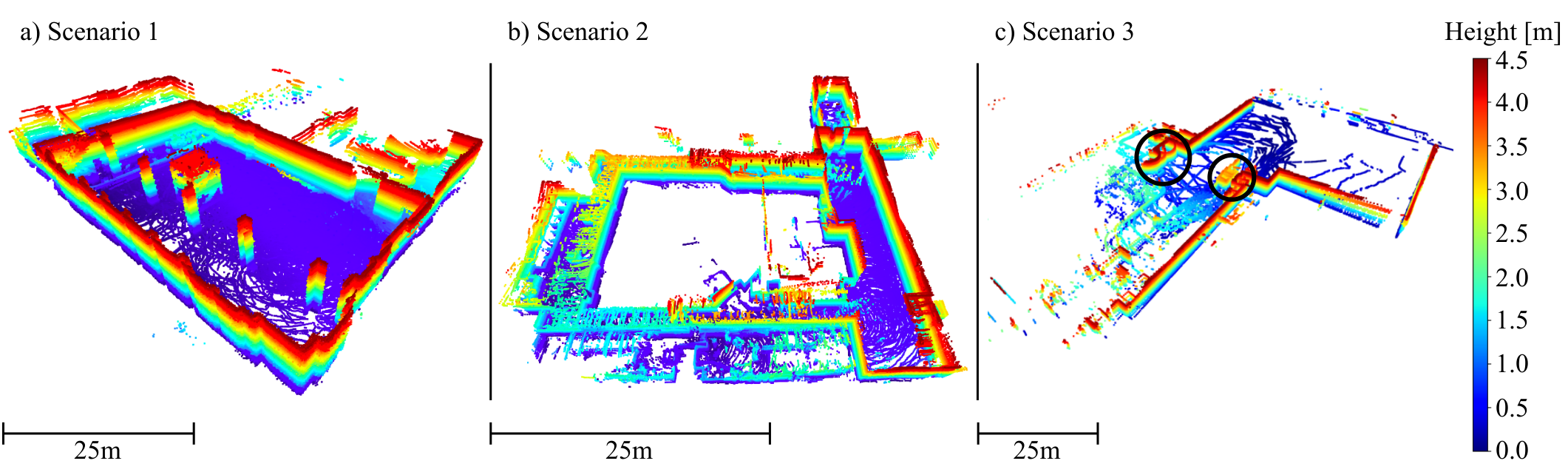}
    \caption{\textbf{Autonomous Mapping results} Visualized are the $3D$ geometry maps of all test scenarios. Color coding is used to indicate map height. In c), map fragmentation caused by challenging outdoor light is marked with black circles.}
    \label{fig:results_3d_plots}
\end{figure*}

\textbf{Initiating Mapping and Navigation}
Autonomous and semi-autonomous operation modes are provided for mapping and navigation.
These are switched using priority-based multiplexers.
Two multiplexers are deployed.

The first multiplexer switches between command velocities generated by the operator and the autonomous navigation stack.
When switching from a signal source with a higher priority to a lower priority, e.g., switching from teleoperation to trajectory planner velocity, switching occurs with a predefined cooldown of \SI{5}{\second} to mitigate robot oscillation.
The second multiplexer switches the goal pose between the following sources in descending priority: (1) "Return" routine, (2) planning preemption, (3) user input, and (4) exploration.
The "Return" routine triggers the robot to return to the home pose in case of losing connection during deployment.
While planning preemption depicts a node that clears trajectory planning if the operator performs teleoperation.
User input refers to the goal pose set by the operator in the $2D$ map of the user interface.
Exploration refers to goal pose specified by the frontier-based exploration~\cite{Hoerner2016}.

\textbf{Operator's Role}
During the autonomous mapping stage, the human operator's role is to monitor the robot's status and to ensure that the mapping process proceeded without errors. 
\ac{ros} RViz provides real-time visualizations of the robot’s environment, trajectory, and the sensor data. 

During field testing it has been shown that manipulating valves requires the operator to refine end-effector poses.
For valve closing the operator specifies a \ac{tcp} pose in the user interface, which is approached using MoveIt~\cite{Moveit2014}.
For substance sampling the operator triggers a movement routine, where the arm picks up the sampling probe from its internal storage and positions it \SI{10}{\cm} over the ground.
In both cases the operator adjusts the robot's pose using a gamepad and the camera's live feed.
Movement routines exist for substance sampling, and substance analysis.
Substances are sampled with a swiping motion over the ground, and based on the routine, the probe is either placed in the tray or immediately analysed using the Raman spectrometer.
To mitigate the risk of self-contamination, the movement routine is designed in such a way that the \ac{tcp} does not pass over the robot body.

\textbf{Complementary Robot Behavior}
\label{sec:complementary}
The "Return" routine triggers the robot to return to the home pose after reaching the goal pose, or when losing connection to the operator.
The robot navigates back to the operator's location, which is set to the location where mapping and exploration started.
Furthermore, to mitigate deadlocks the operator has the ability to override the priority multiplexer.
Manually specifying movement commands overrides autonomous navigation and discards the active goal.

\begin{figure}[t!]
    \centering
    \includegraphics[width=0.75\columnwidth]{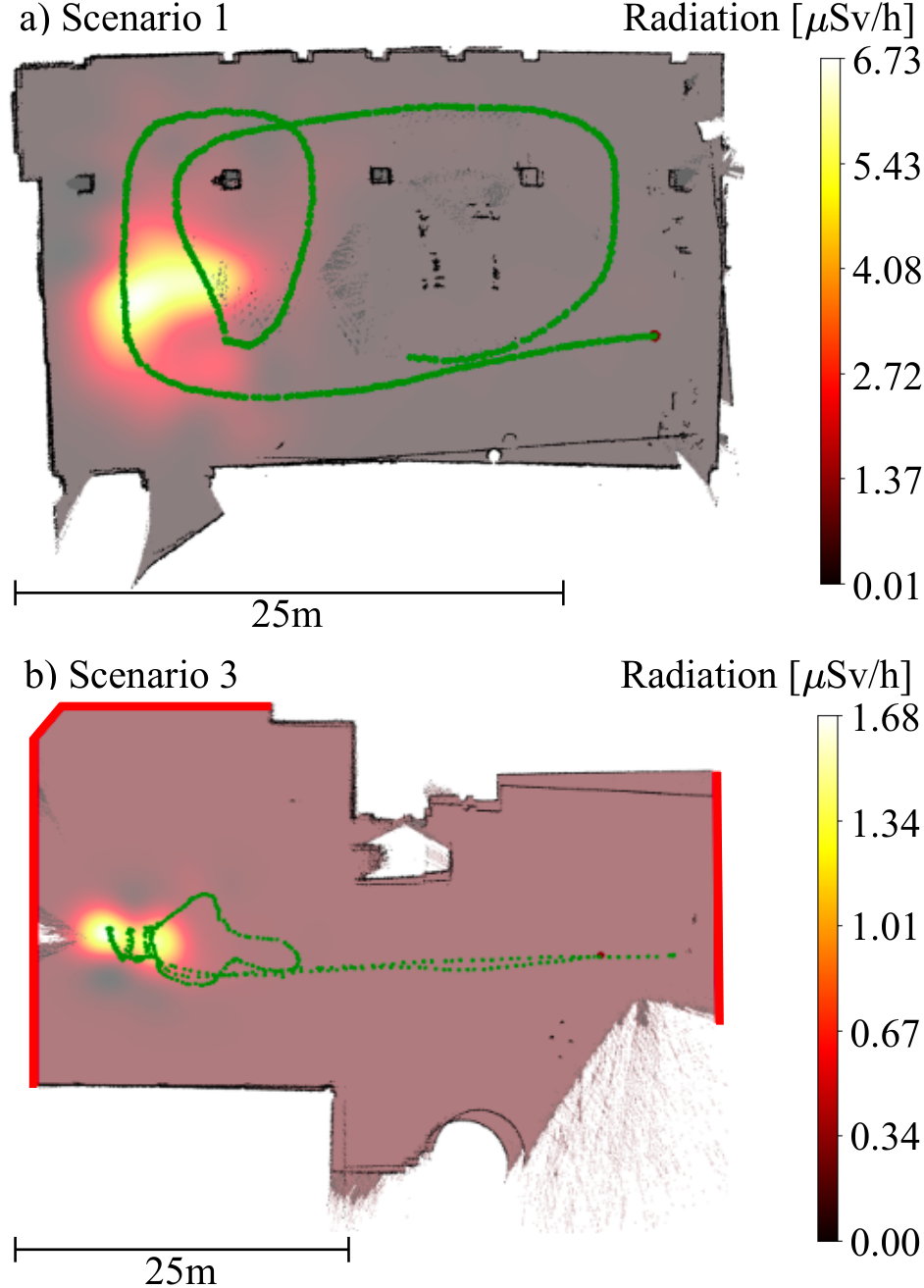}
    \caption{\textbf{Radiation mapping results} 
    $2D$ geometric map (black), overlaid with the robot trajectory (green), and the radiation map (yellow shading). Red borders in b) indicate clipping to highlight relevant map parts.}
    \label{fig:radiation_maps}
\end{figure}

\section{FIELD TRIALS} \label{sec:experiments}

This section presents trials of UGV-CBRN during the \ac{enrich} and at the nuclear, biological, chemical, and disaster relief training area Tritolwerk Eggendorf, Austria.
The \ac{enrich} is hosted at the decommissioned nuclear power plant of Zwentendorf, Austria, and aims to simulate disasters and to test \ac{cbrn} response robots.
UGV-CBRN has been tested in the following three scenarios:

\begin{enumerate}
    \item \textbf{Tritolwerk main production hall}\\
    This scenario features an empty production hall with pillars and scaffolding.
    A radiation source (Cobalt-60) was positioned beneath a pile of bricks to complicate radiation localization, and placed near one of the exits. 
    \item \textbf{Nuclear power plant Zwentendorf (Indoor)}\\%(\ac{enrich} training area)}\\
    This indoor scenario takes places in the corridors of a power plant. 
    The system's capacity to navigate and map narrow hallways and rooms connected by doors was evaluated.
    \item \textbf{Nuclear power plant Zwentendorf (Outdoor)}\\
    This scenario presents a large outdoor area with a human operator navigating the robot toward an unobstructed radiation source (Caesium-137). 
    The robot's trajectory imitates that of a human disaster responders triangulating the radiation source. %The scenario is located outside near the nuclear power plant Zwentendorf.
\end{enumerate}

\subsection{Generated Geometry Maps}

\figref{results_3d_plots} visualizes the $3D$ map generated for all three scenarios. 
The reconstruction of Scenario 1 depicts sharp geometric details, such as the scaffolding, including belt for securing it, in the middle of the room, as well as the barrier at the far entrance of the hall.
Scenario 2 demonstrates a successful loop closure despite occasional navigation of narrow passages with only centimeters in margin.
The turbine rooms (in the left foreground) have not been fully mapped due to the limited spatial capacity within them.
The map of Scenario 3, showing an open large-scale outdoor area, indicates reconstruction deficiencies highlighted with black circles. In these areas, map fragmentation is visible. This happens due to challenging illumination, which results in noisy \ac{lidar} measurements. %

\subsection{Generated Radiation Maps} \label{ch:results_rad_mapping}

\figref{radiation_maps} presents the generated $2D$ map (black) aligned with the radiation map (yellow shading) and overlaid with the robot trajectory (green) of Scenario 1 and Scenario 3. The resulting heatmap represents the triangulated radiation sources.
The heatmap \figref{radiation_maps}a) depicts the shielding of the emitted radiation through the pile of bricks at the bottom right of the radiation source. 
The trajectory in \figref{radiation_maps}b) resembles that of a human responder, circling the anticipated position of the radiation source.
The dumbbell-shaped radiation peak is an artifact of the fragmentation of the geometric map.

\subsection{Substance Sampling Experiments} \label{ch:results_sampling}

\begin{figure}
    \centering
    \vspace{2mm}
    \begin{subfigure}[t]{0.24\columnwidth}
        \centering
        \includegraphics[width=\textwidth]{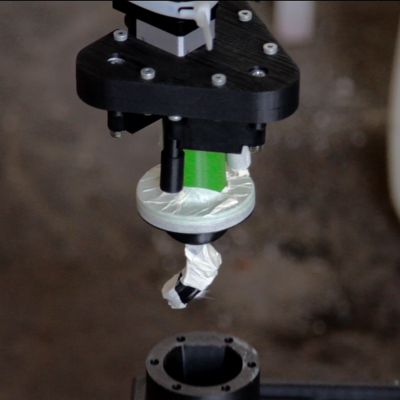}
        \caption{}
    \end{subfigure}
    %\hfill
    \begin{subfigure}[t]{0.24\columnwidth}
        \centering
        \includegraphics[width=\textwidth]{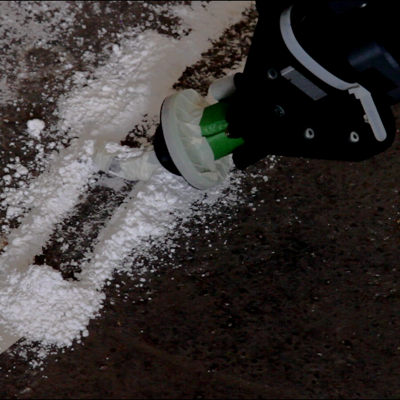}
        \caption{}
    \end{subfigure}
    %\hfill
    %\\ \vspace{1em}
    \begin{subfigure}[t]{0.24\columnwidth}
        \centering
        \includegraphics[width=\textwidth]{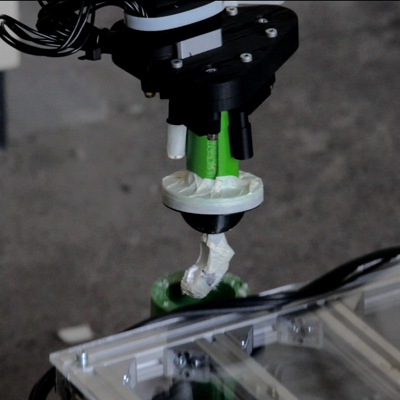}
        \caption{}
    \end{subfigure}
    %\hfill
    \begin{subfigure}[t]{0.24\columnwidth}
        \centering
        \includegraphics[width=\textwidth]{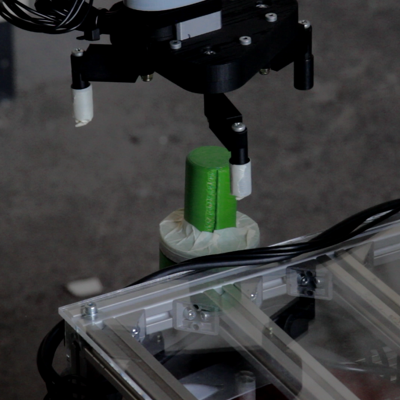}
        \caption{}
    \end{subfigure}
    \caption{\textbf{Semi-autonomous substance sampling process} 
    a) retrieve probe from the storage, b) swipe over the substance, c) position sample over container, and d) sample placed in the analysis tray.}
    \label{fig:results_sampling}
\end{figure}

\figref{results_sampling} visualizes a sampling sequence of a powdered substance, performed in Scenario 1.
Each of the steps is executed with a predefined movement sequence of the arm, each of which is triggered by the human operator in turn. 
After steps a and b the operator refines the end-effector pose via the user interface.
The steps are a) retrieving the probe from the storage and placing it in front of the robot, b) sampling with a swiping motion, c) positioning the sampled substance over the spectrometer without passing it over the robot's body, and d) placing the probe in the analysis tray.

\subsection{Valve Manipulation Experiments}

\begin{figure}[!t]
    \centering
    \vspace{2mm}
    \includegraphics[width=0.4\textwidth]{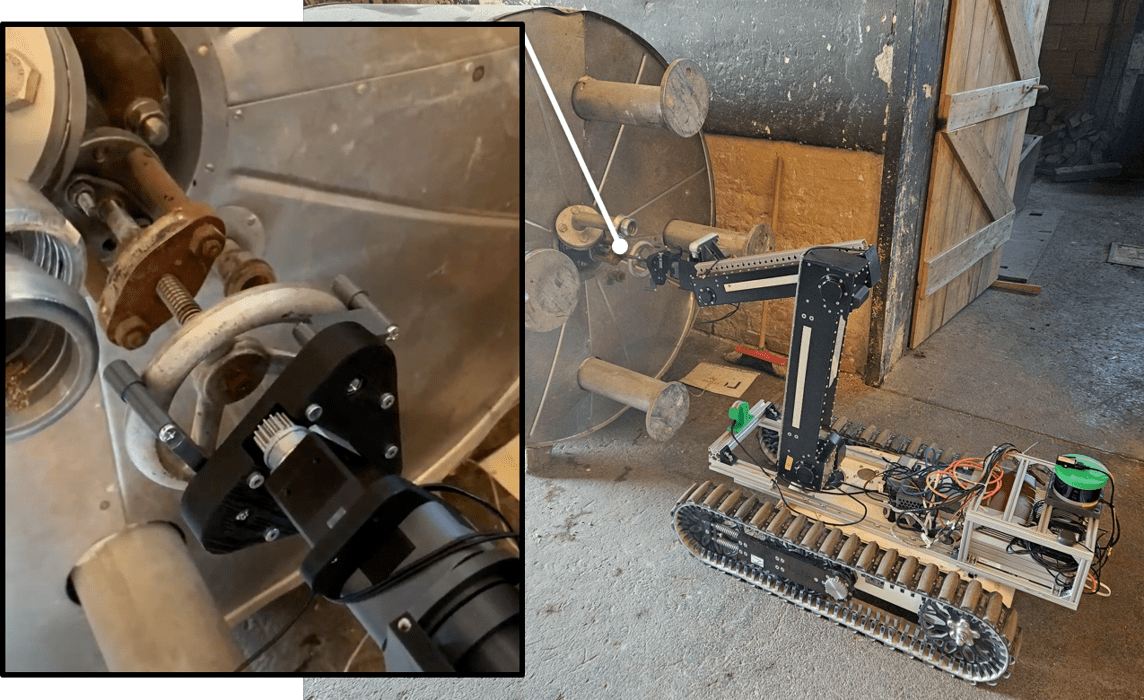}
    \caption{\textbf{Valve manipulation} An external grasp of a $100 mm$ valve in Scenario 1.}
    \label{fig:results_valve_closing}
\end{figure}

\figref{results_valve_closing} visualizes valve closing, performed in Scenario 1. 
A valve with a diameter of approximately \SI{100}{mm} is manipulated with an external grasp. 
The operator positions the end-effector, initiates the gripper closure, and rotates it.
This procedure is performed manually to adequately account for variability in valve designs and environmental conditions.
Although the end-effector is capable of producing sufficient torque, it is a prototype designed to demonstrate the practical functioning without the mechanical material properties for rotating the valve.

\section{CONCLUSION} \label{sec:conclusions}
This manuscript presents an unmanned ground vehicle for chemical, biological, radiological, and nuclear disaster response.
Its capabilities for supporting \ac{sar} include autonomous exploration and mapping of geometry and radiation, sampling and online analysis of encountered substances, as well as the manipulation of valves to stop substance outflows.
The system operation is based on a user interface for monitoring the robot state and a priority-based multiplexer for seamless integration of full- and semi-autonomy. 
Diverse error recovery strategies ensure continuous operation during deployment.
Tests in real-world scenarios with radiation sources showcase the different capabilities of the presented system.

Future work will integrate solutions to improve the robustness of map generation for the challenges of large-scale outdoor environments. 
End-effector design will be re-considered to improve versatility, e.g. for debris removal. 
Furthermore, strategies to improve the autonomy of the robot will be developed and tested. 

\addtolength{\textheight}{-7.cm}   % This command serves to balance the column lengths

{\small
\bibliographystyle{IEEEtran}
\bibliography{root}
}

\end{document}